\documentclass[runningheads]{llncs}
\usepackage[T1]{fontenc}
\usepackage{cite}
\usepackage{amsmath,amssymb,amsfonts}
\usepackage{algorithmic}
\usepackage{graphicx}
\usepackage{textcomp}
\usepackage[dvipsnames,table]{xcolor}
\usepackage{xspace}
\usepackage{adjustbox}
\usepackage{multicol}
\usepackage{multirow}
\usepackage{booktabs}
\usepackage{pifont}
\usepackage{authblk}
\usepackage{colortbl}

\usepackage{graphicx}
\usepackage{floatflt}

\usepackage[compatibility=false]{caption}
\usepackage{subcaption}

\makeatletter
\DeclareRobustCommand\onedot{\futurelet\@let@token\@onedot}
\def\@onedot{\ifx\@let@token.\else.\null\fi\xspace}

\def\eg{\emph{e.g}\onedot}

\def\etal{\emph{et al}\onedot}
\usepackage{hyperref}
\definecolor{cvprblue}{rgb}{0.21,0.49,0.74}
\hypersetup{
    colorlinks=true,
    linkcolor=Blue, 
    citecolor=Blue, 
    urlcolor=purple 
}
\usepackage[capitalize]{cleveref}
\crefname{section}{Sec.}{Secs.}
\Crefname{section}{Section}{Sections}
\Crefname{table}{Table}{Tables}
\crefname{table}{Table}{Tables}
\Crefname{figure}{Figure}{Figures}
\crefname{figure}{Fig.}{Figs.}
\Crefname{equation}{Equation}{Equations}
\crefname{equation}{Eq.}{Eqs.}
\crefname{algocf}{alg.}{algs.}
\Crefname{algocf}{Algorithm}{Algorithms}
\begin{document}
\title{SFormer: SNR-guided Transformer for Underwater Image Enhancement from the Frequency Domain}
%
%
\author{Xin Tian\inst{1} \and
Yingtie Lei\inst{1} \and
Xiujun Zhang\inst{2} \and
Zimeng Li\inst{2}\orcidID{0000-0003-2798-3134}\thanks{Corresponding authors.} \and
Chi-Man Pun\inst{1}\orcidID{0000-0003-1788-3746} \and
Xuhang Chen\inst{1,3}\orcidID{0000-0001-6000-3914}\protect\footnotemark[1]
}
\titlerunning{SFormer}
\authorrunning{Tian et al.}
%
\institute{University of Macau \and
Shenzhen Polytechnic University	\and
School of Computer Science and Engineering, Huizhou University \\
\email{li\_zimeng@szpu.edu.cn, xuhangc@hzu.edu.cn}}
\maketitle              
\begin{abstract}
Recent learning-based underwater image enhancement (UIE) methods have advanced by incorporating physical priors into deep neural networks, particularly using the signal-to-noise ratio (SNR) prior to reduce wavelength-dependent attenuation. However, spatial domain SNR priors have two limitations: (i) they cannot effectively separate cross-channel interference, and (ii) they provide limited help in amplifying informative structures while suppressing noise. To overcome these, we propose using the SNR prior in the frequency domain, decomposing features into amplitude and phase spectra for better channel modulation. We introduce the Fourier Attention SNR-prior Transformer (FAST), combining spectral interactions with SNR cues to highlight key spectral components. Additionally, the Frequency Adaptive Transformer (FAT) bottleneck merges low- and high-frequency branches using a gated attention mechanism to enhance perceptual quality. Embedded in a unified U-shaped architecture, these modules integrate a conventional RGB stream with an SNR-guided branch, forming SFormer. Trained on 4,800 paired images from UIEB, EUVP, and LSUI, SFormer surpasses recent methods with a 3.1 dB gain in PSNR and 0.08 in SSIM, successfully restoring colors, textures, and contrast in underwater scenes.
\end{abstract}

\section{Introduction}
Underwater Image Enhancement (UIE) is essential for underwater exploration and conservation. Light absorption and scattering result in blue or green, low contrast, and blurry underwater images. \Cref{fig:intro} shows the physical processes degrading image quality. These challenges impact marine studies, robotic navigation, environmental monitoring, and photography. Effective UIE methods enhance clarity and detail, improving accuracy in tasks and understanding the environment.
 


\begin{floatingfigure}[l]{0.6\textwidth}
\includegraphics[width=0.55\textwidth]{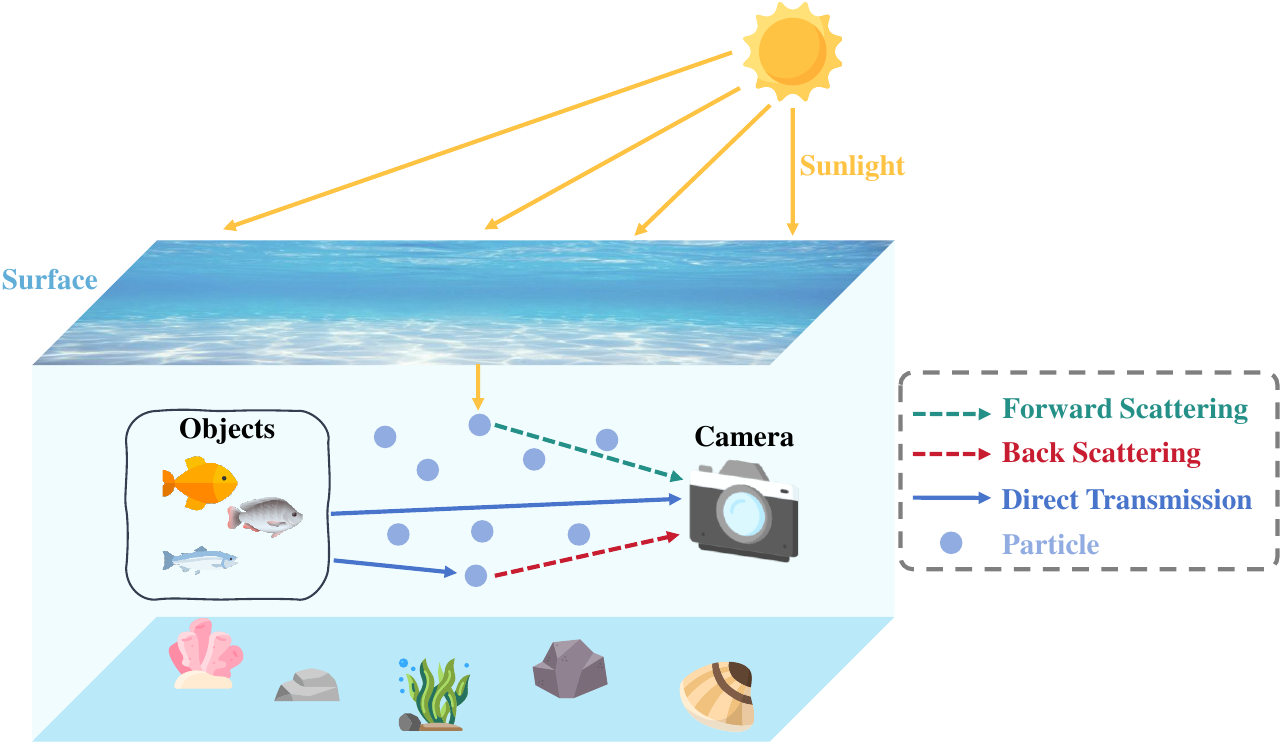}
    \caption{Underwater image formation model. Sunlight entering the water column undergoes attenuation due to absorption and scattering by suspended particles (forward and back scattering), resulting in diminished direct transmission and characteristic underwater image artifacts.}
    \label{fig:intro}
\end{floatingfigure}

While traditional UIE methods have made progress, they are inconsistent in complex water conditions. Recently, deep learning, notably CNNs, has become a strong alternative due to its ability to extract local features~\cite{huang2023contrastive,li2020underwater,jiang2020geometry,li2022few,liu2023coordfill,li2021hybrid,li2022monocular,zhang2022correction,wang2023qgd,wang2024beyond}. However, CNNs struggle with global context~\cite{jiang2022underwater}. This has led to exploring Transformer-based architectures~\cite{peng2023u,lei2024uie,chen2024uwformer,lei2025cmamrnet,liu2024forgeryttt,liu2024depth,zheng2024lagrange,li2025gre}, which are effective in capturing long-range dependencies and context.

Despite advances in network architectures, most UIE methods fail to address a key issue in underwater environments: SNR degradation due to light attenuation and scattering~\cite{akkaynak2018revised}. Underwater scenarios face wavelength-dependent absorption and scattering, causing spatially varying SNR degradation across color channels and image regions~\cite{schechner2005recovery}. Conventional methods using CNNs or Transformers process RGB signals without accounting for photon-statistical distortions, thus limiting achievable restoration quality.

Previous studies have integrated SNR priors into deep learning mostly in the spatial domain, which limits their flexibility and global feature adjustment in underwater conditions~\cite{xu2022snr}. Frequency-domain methods using FFT can decompose images into low- and high-frequency components and separate amplitude and phase~\cite{wang2023ultra}, aiding noise suppression and frequency enhancement. Recent research indicates that frequency-domain priors in deep learning enhance models by capturing features missed by spatial methods~\cite{miao2021underwater,le2024medical,xu2024plaintext,li2025dppad,zhong2025image}. We propose using SNR priors in the frequency domain to restore amplitude and phase, improving underwater image enhancement.

We propose a novel SNR-guided transformer (SFormer) UIE framework with a U-shaped architecture of CNN blocks for hierarchical feature extraction. At the bottleneck, the Frequency Adaptive Transformer (FAT) module models long-range dependencies and frequency domain features, capturing global context and enhancing frequency characteristics. To combat SNR degradation in the frequency domain, we introduce the Fourier Attention SNR-prior Transformer (FAST), combining Fourier analysis and SNR-informed attention to emphasize important spectral components and reduce noise artifacts. Key contributions are summarized as follows:
\begin{enumerate}
    \item We propose the SFormer UIE framework, a novel U-shaped frequency-aware architecture guided by SNR priors, which incorporates the FAST and FAT modules. Extensive experimental results validate that our framework consistently outperforms current state-of-the-art methods in UIE tasks.
    \item We introduce the Fourier Attention SNR-prior Transformer (FAST) module, an advanced attention mechanism leveraging amplitude-phase interactions in the frequency domain, explicitly conditioned on SNR priors, thereby significantly enhancing frequency-dependent feature extraction.
    \item We develop the Frequency Adaptive Transformer (FAT) module, designed to adaptively fuse low- and high-frequency information through channel and spatial attention mechanisms. Further refined by an effective gating strategy, this module optimizes information fusion, substantially enhancing overall image enhancement performance.
\end{enumerate}

\section{Related Work}
\subsection{Traditional Underwater Image Enhancement}
Traditional UIE methods can be broadly categorized into two main groups: physical model-free methods and physical model-based methods~\cite{guo2025underwater,wu2025image,wu2025generative}.

Physical model-free methods enhance visual quality by directly manipulating image properties without incorporating physical underwater imaging models. These approaches employ conventional image processing techniques such as histogram equalization~\cite{ghani2017automatic}, contrast enhancement~\cite{ancuti2012enhancing}, and white balance correction~\cite{ancuti2017color} to modify pixel values. Although computationally efficient and straightforward to implement, these methods typically fail to address the complex physical mechanisms that underlie underwater image degradation.

In contrast, physical model-based methods leverage environmental priors and predefined assumptions to estimate parameters such as transmission maps~\cite{drews2013transmission,wu2024top,wuimgfu,wu2025prompt}, background light~\cite{zhao2015deriving}, and wavelength-dependent attenuation coefficients~\cite{akkaynak2019sea}. By formulating UIE as an inverse problem, these approaches systematically attempt to reconstruct original scenes by reversing degradation effects according to established physical imaging models. Notable examples include techniques using the Dark Channel Prior (DCP)~\cite{he2010single}, Underwater Light Attenuation Prior (ULAP)~\cite{zhou2023underwater}, and statistical priors~\cite{peng2017underwater}. Despite plausible theoretical foundations, their dependence on rigid physical assumptions inherently restricts their effectiveness across the diverse spectrum of underwater imaging conditions.

\subsection{Learning-based Underwater Image Enhancement}
Learning-based UIE methods effectively address the limitations inherent in conventional approaches via deep learning~\cite{liu2023explicit,li2023cee,zhu2024test,li2024cross,zheng2024smaformer,li2025adaptive,liu2024dh,wang2024novel,wang2024novel2,wang2025structure}. These frameworks leverage the representational power of deep neural networks to automatically learn hierarchical features (\eg, color distributions, texture patterns) from diverse underwater scenarios. The data-driven nature enables them to handle environmental variability and optical distortions with greater flexibility and accuracy. 

Early learning methods relying on physical models struggled in complex conditions, while CNN-based solutions introduced end-to-end pipelines for color correction and detail enhancement. However, CNN models including WaterNet~\cite{li2019underwater}, Ucolor~\cite{li2021underwater}, and UWCNN~\cite{li2020underwater} require large, high-quality datasets for training. To reduce reliance on large datasets, GAN-based strategies are introduced, like CycleGAN~\cite{zhu2017unpaired}, UGAN~\cite{fabbri2018enhancing} and PUGAN~\cite{cong2023pugan}. Although they show potential, GANs can suffer from model collapse and training instability problems. More recently, Transformer-based networks have shown promise in capturing long-range dependencies. For example, architectures like the U-shaped Transformer~\cite{peng2023u} have been introduced to enhance the efficiency and accuracy of model learning.

In addition, there has been growing interest in integrating imaging priors into deep learning models to improve adaptability under different water conditions. Ucolor~\cite{li2021underwater} uses a medium transmission-guided multicolor space embedding to address color casts and low contrast due to wavelength and attenuation. Mu ~\etal~\cite{mu2023generalized} proposed a dynamic method guided by physical knowledge for adaptive enhancement. Although these approaches have advanced the field~\cite{zhang1,zhang2,zhang3,zhang4,zhang5,zhang6,zhang7,zhang8,zhang9,zhang10,zhang11,zhang12}, no existing method has yet explored the use of SNR priors in the frequency domain. Our work aims to fill this gap by investigating how frequency-domain SNR priors can further improve UIE performance.

\section{Methodology}

\subsection{Overall architecture}
\begin{figure}[ht]
  \centering
  \includegraphics[width=\textwidth]{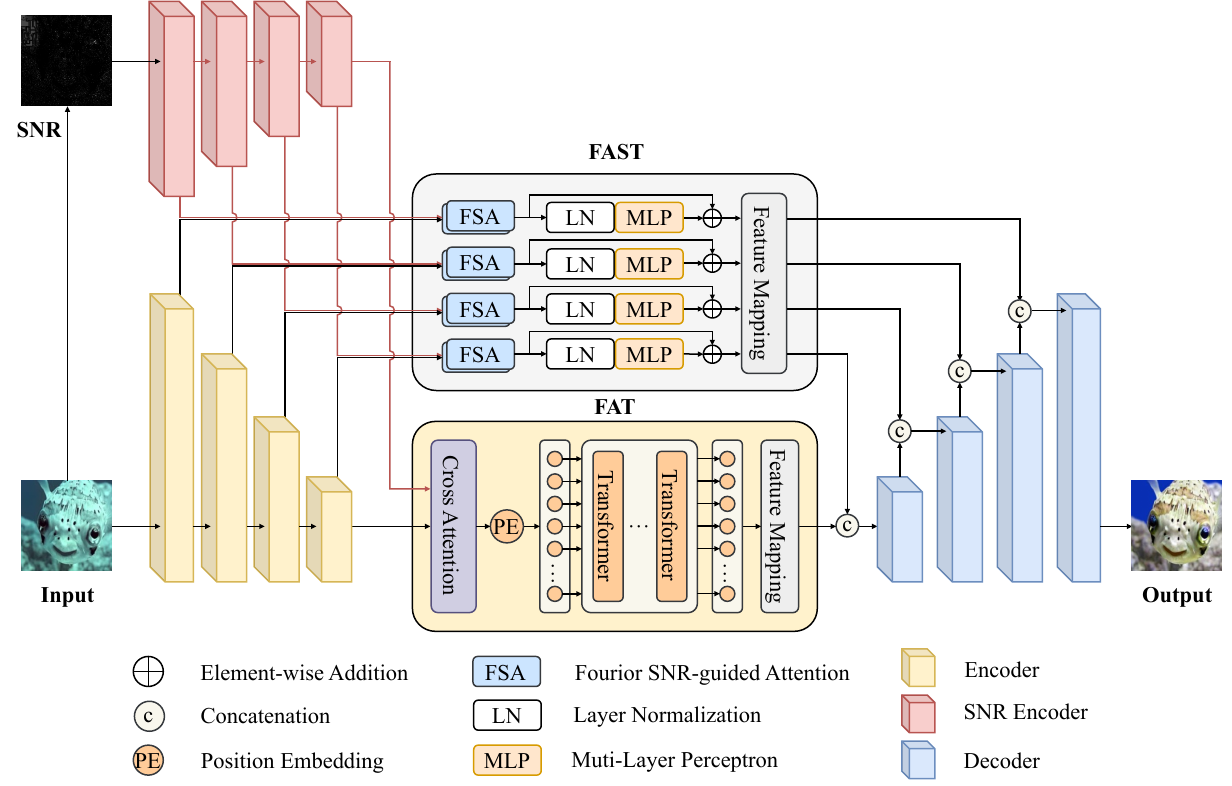}
  \caption{Overview of SFormer architecture, a dual-branch U-shaped structure integrating RGB images and SNR priors.
  }
  \label{fig:uie}
\end{figure}

The overall architecture, illustrated in \cref{fig:uie}, comprises two main modules: Frequency-Adaptive Transformer (FAT) and Fourier Attention SNR-prior Transformer (FAST).

Our proposed network adopts a dual-branch U-shaped design, integrating a conventional RGB encoder with an SNR-prior branch. These branches converge at a transformer-based bottleneck and subsequently reunite through a symmetric decoder. Specifically, the RGB encoder processes the input image $I_{input} \in \mathbb{R}^{3 \times H \times W}$ through four consecutive convolutional blocks. Each block consists of two convolutional layers with $3 \times 3$ kernels and ReLU activations, followed by $2 \times 2$ max-pooling operations. After the fourth pooling operation, the spatial resolution is reduced to $\frac{H}{16} \times \frac{W}{16}$, generating multi-scale feature maps.

Simultaneously, a single-channel SNR map $S\in \mathbb{R}^{1 \times H \times W}$ is derived from the input image. This SNR map undergoes an identical four-stage down-sampling procedure, resulting in feature representations aligned both spatially and semantically with the RGB features. The RGB and SNR streams remain orthogonal at this stage, interacting exclusively through FAST modules positioned at each encoder skip connection and the FAT bottleneck at the deepest layer. The specific configurations and computations within FAST and FAT are detailed in the subsequent subsections.

The decoder mirrors the structure of the encoder. The FAT output feature map undergoes $2 \times 2$ transposed convolutional upsampling, and is concatenated with corresponding FAST-enhanced skip feature maps, which are precisely center-cropped for alignment. Each concatenation is processed by convolutional blocks to reconstruct local textures lost during down-sampling. After four consecutive upsampling stages, the original spatial resolution is fully restored. Finally, a $1 \times 1$ convolutional layer predicts an RGB residual map that, when combined with the original input, generates the enhanced output image.

\subsection{Frequency-Aware SNR-adaptive Transformer}
To leverage the complementary advantages of spatial and frequency domains, we propose the FAST module, operating directly within the frequency domain to enhance structural coherence and exploit frequency-aware attention.

Within each FAST module, the RGB feature maps $x_i$ and their corresponding SNR maps $s_i$ are independently projected into query ($Q$), key ($K$), and value ($V$) tensors. Subsequently, the $Q$ and $K$ tensors undergo Fast Fourier Transform (FFT), decomposing into amplitude and phase spectra:
\begin{align}
A_x &= \text{Amp}(\mathcal{F}(K)), \quad P_x = \text{Pha}(\mathcal{F}(K)),\\
A_s &= \text{Amp}(\mathcal{F}(Q)), \quad P_s = \text{Pha}(\mathcal{F}(Q)),
\end{align}
where $\mathcal{F}$ denotes the FFT operator. The amplitude and phase spectra are multiplied element-wise separately to model structural correlations between RGB and SNR features effectively. This procedure highlights mutual frequency characteristics, enabling an enhanced structural representation. The combined frequency representation is transformed back into the spatial domain by inverse FFT (IFFT) as follows: 
\begin{align}
F &= \mathcal{F}^{-1}(A_x \cdot A_s, P_x \cdot P_s).
\end{align}



\begin{floatingfigure}[l]{0.6\textwidth}
\includegraphics[width=0.55\textwidth]{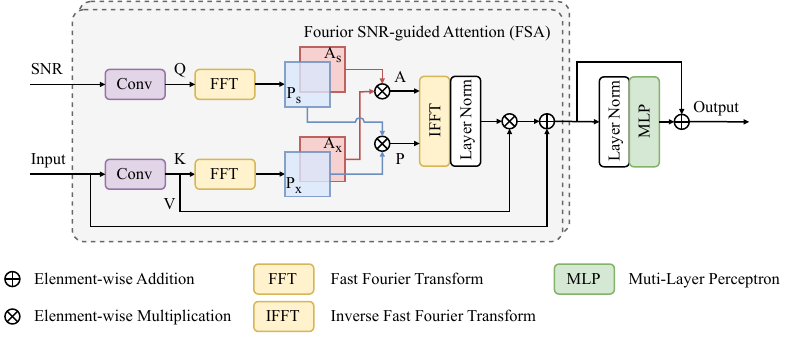}
    \caption{Detailed structure of FAST module.}
  \label{fig:fast}
\end{floatingfigure}

This frequency-aware attention $F$ output is subsequently modulated through element-wise multiplication with the value tensor $V$. A residual connection is then established between the attention-enhanced output and the initial RGB features, preserving essential low-frequency content and facilitating training stability. Additionally, layer normalization (LN) and a lightweight multilayer perceptron (MLP) refine the output to enhance convergence:
\begin{align}
x'_{i} &= x_{i} + \text{LN}(F)\cdot V,\\
x_{out} &= x'_{i} + \text{MLP}(\text{LN}(x'_{i})).
\end{align}
This structure effectively integrates SNR-guided frequency attention, preserving spatial fidelity and reinforcing structurally significant features.

\subsection{Frequency Adaptive Transformer}

The FAT module replaces the traditional bottleneck to serve as a global contextual aggregation hub, emphasizing severely degraded image regions through selective frequency modulation. Recognizing that not all frequency components equally contribute to image enhancement, FAT explicitly decomposes input features into low-frequency (holistic contexts such as illumination) and high-frequency components (localized details and edges), subsequently refining each through specialized attention mechanisms.



\begin{floatingfigure}[l]{0.6\textwidth}
\includegraphics[width=0.55\textwidth]{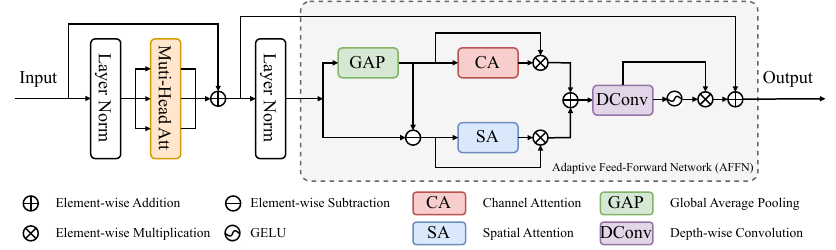}
    \caption{The depiction of our FAT module.}
    \label{fig:fat}
\end{floatingfigure}

Given an input feature map $M \in \mathbb{R}^{C \times \frac{H}{16} \times \frac{W}{16}}$, it is partitioned into non-overlapping $pH \times pW$ patches by a $4 \times 4$ convolution with matching stride. Each patch is flattened and projected into embedding sequences $T_R$. Similarly, the co-located SNR map is projected into $T_S$. Cross-attention then integrates SNR priors by treating $T_R$ as queries and $T_S$ as keys and values, supplemented with positional embeddings:
\begin{align}
P &=\text{CrossAtt}(T_R, T_S) + T_R + \text{PE},
\end{align}
where PE represents a position embedding operation. The resulting patch $P$ passes through transformer blocks, each comprising multi-head self-attention (MHA) and adaptive feed-forward network (AFFN). Formally, the output of the $i^{th}$ transformer layer is defined as:
\begin{align}
P'_i &= \text{}\text{MHA}(\text{LN}(P_{i-1})) + P_{i-1}, \\
P_i &= \text{AFFN}(\text{LN}(P'_i)) + P'_i.
\end{align}

The AFFN converts tokens back to spatial maps, separating them into low-frequency $F_{low}$ and high-frequency $F_{high}$ components as follows:
\begin{align}
F_{low} &= \text{GAP}(\text{LN}(P'_i)), \\ 
F_{high} &= \text{LN}(P'_i) - F_{low},
\end{align}
Then Channel-wise attention (CA) enhances low-frequency components, optimizing global contrast and color balance, while spatial attention (SA) accentuates critical edges in high-frequency domain. These two branches are then merged by element-wise addition, combining holistic and localized representations:
\begin{align}
F' &= \text{CA}(F_{low}) \cdot F_{low} + \text{SA}(F_{high}) \cdot F_{high},
\end{align}
Subsequent depth-wise convolution further integrates these features, followed by a gated GELU activation to selectively reinforce significant details as:
\begin{align}
F'_1, F'_2 &= \text{Chunk}(\text{DConv}(F')),\\
P_i &= P'_i + \text{GELU}(F'_1) \cdot F'_2.
\end{align}
This gating mechanism discards redundant frequency components, retaining only the most informative features. By explicitly preserving frequency discrimination within each transformer layer, AFFN effectively addresses the over-smoothing inherent in conventional Vision Transformer, promoting robust detail restoration and enhanced semantic accuracy.

\subsection{Loss Function}

To ensure both pixel-level fidelity and perceptual realism, we optimize the network using a composite loss that simultaneously constrains the spatial, frequency, and color domains. Specifically, histogram-aware losses are employed in the perceptually uniform $LAB$ and $LCH$ color spaces. These spaces effectively separate luminance from chromatic information, enabling color correction without compromising local luminance contrast.

In the $LAB$ loss function, the luminance channel $L$ is penalized using $\mathcal{L}_2$ loss, while the chromatic channels $A$ and $B$ employ soft cross-entropy losses as:
\begin{align}
\mathcal{L}_{LAB}(pre, gt) 
&= \mathbb{E}_{in, gt} \Bigg[ (L^{gt} - L^{pre})^2  \nonumber - \sum_{i=1}^{n} Q(A_i^{gt}) \log(Q(A_i^{pre})) \nonumber \\
&\quad - \sum_{i=1}^{n} Q(B_i^{gt}) \log(Q(B_i^{pre})) \Bigg],
\end{align}
where $Q$ stands for the quantization operator. This approach minimizes chromatic discrepancies between the predicted and ground truth images. Similarly, the $LCH$ loss function targets the chroma $C$ and hue $H$ channels explicitly. The chroma discrepancies and hue deviations are penalized by $\mathcal{L}_2$ loss, while luminance channel $L$ is addressed by applying a soft cross-entropy as:
\begin{align}
\mathcal{L}_{LCH}(pre, gt) 
&= \mathbb{E}_{in, gt} \Bigg[ 
- \sum_{i=1}^{n} Q(L_i^{gt}) \log(Q(L_i^{pre})) \nonumber \\
&\quad + (C^{gt} - C^{pre})^2 + (H^{gt} - H^{pre})^2 
\Bigg].
\end{align}

The spatial domain $\mathcal{L}_1$ loss is utilized to enforce precise pixel-level reconstruction, and a frequency domain loss, calculated using the Fourier transform, is incorporated to maintain structural similarity in the frequency spectrum and stabilize gradient propagation.
\begin{align}
\mathcal{L}_s &= \frac{1}{T} \left\| pre - gt \right\|_1, \\
\mathcal{L}_f &= \frac{1}{T} \left\| \mathcal{F}(pre) - \mathcal{F}(gt) \right\|_1,
\end{align}
where $T$ represents the total elements. In addition, a perceptual loss $\mathcal{L}_{per}$ using the Learned Perceptual Image Patch Similarity metric (LPIPS)~\cite{panetta2015human}, based on a VGG network, ensures semantic fidelity and visual realism. Then the total loss function combines these individual components into a comprehensive objective:
\begin{align}
\mathcal{L}_t = \alpha \, \mathcal{L}_s + \beta \, \mathcal{L}_f + \gamma \, \mathcal{L}_{LAB} + \mu \, \mathcal{L}_{LCH} + \lambda \, \mathcal{L}_{per}.
\end{align}
Here, the hyperparameters $(\alpha, \beta, \gamma, \mu, \lambda)$ are empirically set as (100, 10, 0.0001, 1, 100), respectively.

\section{Experiment}

\subsection{Experimental Datasets}
We assess our underwater image enhancement model on three benchmark datasets: UIEB~\cite{li2019underwater}, EUVP~\cite{islam2020fast}, and LSUI~\cite{peng2023u}. We create training sets of 800 images from UIEB, 2000 from EUVP, and 2000 from LSUI for a balanced dataset. Test sets include 90 images from UIEB, 200 from EUVP, and 200 from LSUI, following standard splits with no overlap. We also retrain state-of-the-art models with these training sets for fair comparison.

\subsection{Implementation Details}
Our model uses PyTorch and trains on a NVIDIA V100 GPU. We use AdamW optimizer with an initial learning rate of $1e-4$, reduced to $1e-6$ via cosine annealing with periodic restarts to avoid local minima and improve convergence. Training lasts 300 epochs with a batch size of 4. Images are resized to $256 \times 256$ pixels. We employ data augmentation techniques like random cropping, flipping, rotation, scaling, and mixup for robustness.

\subsection{Evaluation Metrics}
To evaluate our method, we use four full-reference metrics: PSNR, SSIM, LPIPS~\cite{zhang2018unreasonable}, and delta error in perceptual color difference ($\Delta E$), plus one non-reference metric, UCIQE~\cite{yang2015underwater}. PSNR measures pixel-level fidelity; higher values mean better accuracy. SSIM assesses perceptual similarity by considering luminance, contrast, and structure; higher scores show better visual preservation. LPIPS uses neural networks to evaluate perceptual similarity, with lower scores indicating better results. $\Delta E$ assesses color accuracy differences; lower values mean minimal perceptual discrepancies. UCIQE evaluates underwater image quality based on human visual perception; higher scores indicate better quality.

\begin{table}[ht]
    \centering
    \caption{Quantitative results on EUVP, LSUI, UIEB datasets. The best is marked in bold and red background. The second-best and third-best are separately marked in blue and green background.}
    \label{tab:result}
    \begin{adjustbox}{width=\linewidth}{
    \begin{tabular}{l|ccccc|ccccc|ccccc}
        \toprule
        \multirow{2}{*}{Methods} & \multicolumn{5}{c|}{EUVP} & \multicolumn{5}{c|}{LSUI} & \multicolumn{5}{c}{UIEB} \\
        \cmidrule(lr){2-6} \cmidrule(lr){7-11} \cmidrule(lr){12-16}
        & PSNR\textuparrow & SSIM\textuparrow & LPIPS\textdownarrow & UCIQE\textuparrow & $\Delta$ E\textdownarrow & PSNR\textuparrow & SSIM\textuparrow & LPIPS\textdownarrow & UCIQE\textuparrow & $\Delta$ E\textdownarrow & PSNR\textuparrow & SSIM\textuparrow & LPIPS\textdownarrow & UCIQE\textuparrow & $\Delta$ E\textdownarrow \\
        \midrule

        Ucolor\cite{li2021underwater} & 23.814 & 0.914 & 0.254 & 0.821 & 7.891 & 21.297 & 0.821 & 0.225 & 0.692 & 12.044 &18.374 & 0.814 & 0.221 & 0.591 & 15.933\\
        CLUIE-Net\cite{li2022beyond} & 24.975 & 0.923 & 0.236 & 0.848 & 6.818 & 23.571 & 0.864 & 0.175 & 0.744 & 9.568 & 19.948 & 0.874 & 0.168 & 0.664 & 14.200\\
        STSC\cite{wang2022semantic} & 25.204 & 0.923 & \cellcolor{YellowGreen!18}0.227 & 0.862 & 6.930 & 24.264 & 0.851 & 0.181 & 0.751 & 8.550 & 21.203 & 0.820 & 0.183 & 0.697 & 12.252\\
        TACL\cite{liu2022twin} & 21.030 & 0.891 & 0.314 & 0.839 & 11.262 & 22.972 & 0.828 & 0.176 & 0.757 & 10.529 & 19.831 & 0.761 & 0.222 & 0.680 & 13.028\\
        UIE-WD\cite{ma2022wavelet} & 17.827 & 0.881 & 0.309 & 0.825 & 14.010 & 19.233 & 0.803 & 0.284 & 0.713 & 15.301 & 20.275 & 0.848 & 0.198 & 0.646 & 14.941\\
        URSCT\cite{ren2022reinforced} & \cellcolor{YellowGreen!18}25.888 & 0.929 & 0.227 & 0.857 & \cellcolor{YellowGreen!18}6.118 &\cellcolor{SkyBlue!18}25.867 & \cellcolor{SkyBlue!18}0.883 & \cellcolor{SkyBlue!18}0.146 & 0.766 & \cellcolor{YellowGreen!18}7.590 &\cellcolor{YellowGreen!18}22.773 & \cellcolor{YellowGreen!18}0.915 & 0.120 & 0.711 & 11.866 \\
        USUIR\cite{fu2022unsupervised} & 22.017 & 0.907 & 0.277 & 0.854 & 10.470 & 23.747 & 0.860 & 0.184 & 0.771 & 10.015 & 22.484 & 0.907 & 0.124 & \cellcolor{pink!18}\textbf{0.724} & \cellcolor{YellowGreen!18}11.312\\
        GUPDM\cite{mu2023generalized} & 24.914 & \cellcolor{YellowGreen!18}0.925 & 0.238 & \cellcolor{SkyBlue!18}0.874 & 7.268 & 25.325 & 0.877 & \cellcolor{YellowGreen!18}0.150 & 0.765 & 8.244 & 22.133 & 0.903 & 0.131 & 0.683 & 12.515\\
        PUGAN\cite{cong2023pugan} & 22.664 & 0.912 & 0.264 & \cellcolor{pink!18}\textbf{0.886} & 8.621 & 23.135 & 0.836 & 0.216 & \cellcolor{SkyBlue!18}0.777 & 9.551 & 20.524 & 0.812 & 0.216 & 0.716 & 12.859\\
        Semi-UIR\cite{huang2023contrastive} & 24.698 & 0.911 & 0.249 & 0.861 & 6.904 & 25.397 & 0.843 & 0.160 & 0.763 & 7.843 & \cellcolor{SkyBlue!18}23.644 & 0.888 & 0.120 & 0.693 & 10.663\\
        SyreaNet\cite{wen2023syreanet} & 23.336 & 0.919 & 0.277 & 0.854 & 10.563 & 24.905 & 0.872 & 0.168 & 0.764 & 9.534 & 22.721 & \cellcolor{SkyBlue!18}0.918 & \cellcolor{SkyBlue!18}0.116 & \cellcolor{SkyBlue!18}0.723 & 11.549\\
        TUDA\cite{wang2023domain} & 23.830 & 0.922 & 0.274 & 0.858 & 9.824 & \cellcolor{YellowGreen!18}25.515 & \cellcolor{YellowGreen!18}0.878 & 0.154 & \cellcolor{YellowGreen!18}0.773 & 9.008 & 22.719 & \cellcolor{YellowGreen!18}0.915 & \cellcolor{YellowGreen!18}0.118 & \cellcolor{SkyBlue!18}0.723 & 11.391\\
        U-Transformer\cite{peng2023u} & 25.118 & 0.913 & 0.274 & 0.859 & 6.642 & 25.151 & 0.838 & 0.221 & 0.765 & 7.801 & 20.747 & 0.810 & 0.228 & \cellcolor{YellowGreen!18}0.715 & 12.719\\
        SGNet\cite{zhao2024toward} & \cellcolor{SkyBlue!18}26.192 & \cellcolor{SkyBlue!18}0.933 & \cellcolor{SkyBlue!18}0.214 & 0.855 & \cellcolor{SkyBlue!18}6.054 & 24.004 & 0.868 & 0.239 & 0.764 & \cellcolor{SkyBlue!18}7.021 & 19.573 & 0.685 & 0.214 & 0.705 & \cellcolor{SkyBlue!18}11.214\\

        Ours & \cellcolor{pink!18}\textbf{27.451} & \cellcolor{pink!18}\textbf{0.936} & \cellcolor{pink!18}\textbf{0.163} & \cellcolor{YellowGreen!18}0.863 & \cellcolor{pink!18}\textbf{4.952}
              & \cellcolor{pink!18}\textbf{26.946} & \cellcolor{pink!18}\textbf{0.956} & \cellcolor{pink!18}\textbf{0.136} & \cellcolor{pink!18}\textbf{0.780} & \cellcolor{pink!18}\textbf{6.007}
              & \cellcolor{pink!18}\textbf{23.680} & \cellcolor{pink!18}\textbf{0.964} & \cellcolor{pink!18}\textbf{0.099} & \cellcolor{pink!18}\textbf{0.724} & \cellcolor{pink!18}\textbf{8.027}\\
        \bottomrule
    \end{tabular}
    }
    \end{adjustbox}
\end{table}

\subsection{Experimental Results and Analysis}
\Cref{tab:result} quantitatively compares the performance of our proposed method against state-of-the-art UIE techniques. The results demonstrate that our method consistently achieves superior performance across all evaluation metrics. Specifically, compared to SGNet~\cite{zhao2024toward}, our method exceeds 2.770 dB of PSNR. Compared to URSCT, the most competitive baseline, our approach achieves significant gains, with average improvements of 1.183 dB in PSNR and 0.043 in SSIM. In addition, notable improvements are observed in other critical metrics, including substantial reductions in LPIPS and $\Delta E$ values, indicating perceptually better color accuracy and visual quality.

\begin{figure}[ht]
  \centering %

  \begin{subfigure}[b]{0.119\linewidth}
    \includegraphics[width=\linewidth]{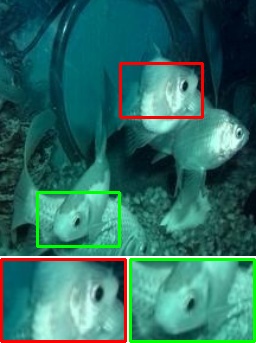}
  \end{subfigure}\hspace{2pt}%
  \begin{subfigure}[b]{0.119\linewidth}
    \includegraphics[width=\linewidth]{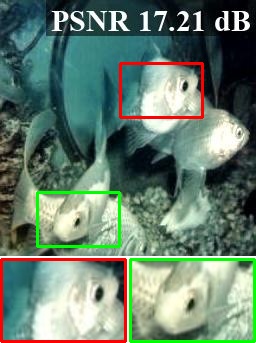}
  \end{subfigure}\hspace{2pt}%
  \begin{subfigure}[b]{0.119\linewidth}
    \includegraphics[width=\linewidth]{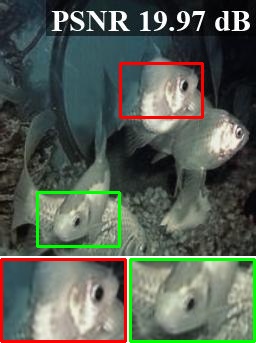}
  \end{subfigure}\hspace{2pt}%
  \begin{subfigure}[b]{0.119\linewidth}
    \includegraphics[width=\linewidth]{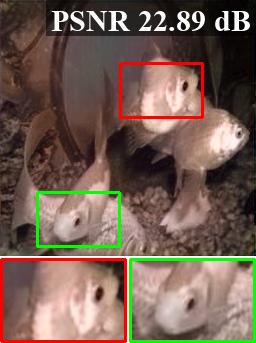}
  \end{subfigure}\hspace{2pt}%
  \begin{subfigure}[b]{0.119\linewidth}
    \includegraphics[width=\linewidth]{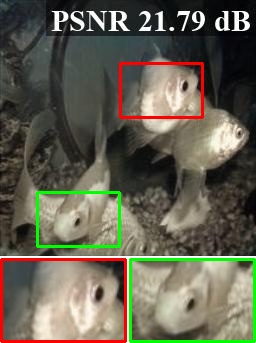}
  \end{subfigure}\hspace{2pt}%
  \begin{subfigure}[b]{0.119\linewidth}
    \includegraphics[width=\linewidth]{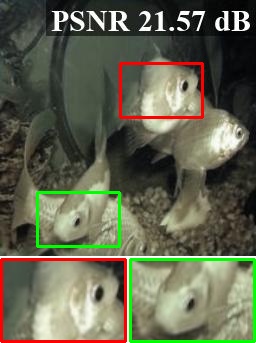}
  \end{subfigure}\hspace{2pt}%
  \begin{subfigure}[b]{0.119\linewidth}
    \includegraphics[width=\linewidth]{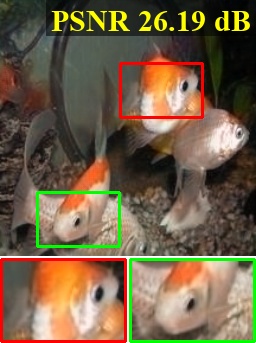}
  \end{subfigure}\hspace{2pt}%
  \begin{subfigure}[b]{0.119\linewidth}
    \includegraphics[width=\linewidth]{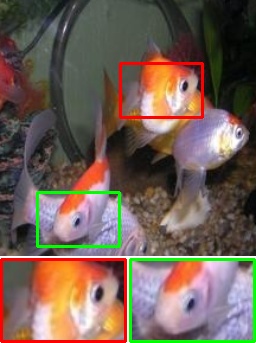}
  \end{subfigure}

  \begin{subfigure}[b]{0.119\linewidth}
    \includegraphics[width=\linewidth]{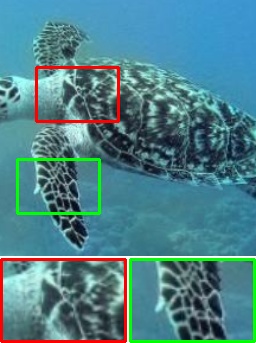}
  \end{subfigure}\hspace{2pt}%
  \begin{subfigure}[b]{0.119\linewidth}
    \includegraphics[width=\linewidth]{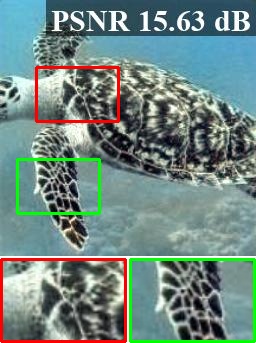}
  \end{subfigure}\hspace{2pt}%
  \begin{subfigure}[b]{0.119\linewidth}
    \includegraphics[width=\linewidth]{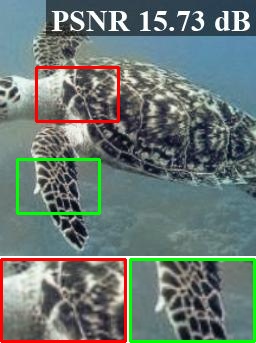}
  \end{subfigure}\hspace{2pt}%
  \begin{subfigure}[b]{0.119\linewidth}
    \includegraphics[width=\linewidth]{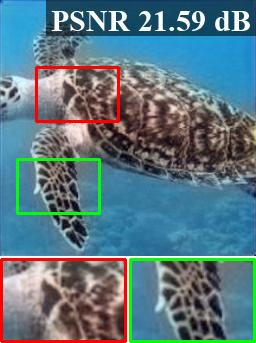}
  \end{subfigure}\hspace{2pt}%
  \begin{subfigure}[b]{0.119\linewidth}
    \includegraphics[width=\linewidth]{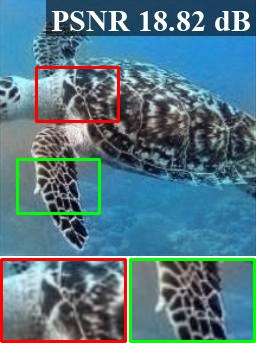}
  \end{subfigure}\hspace{2pt}%
  \begin{subfigure}[b]{0.119\linewidth}
    \includegraphics[width=\linewidth]{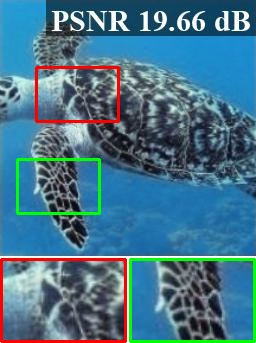}
  \end{subfigure}\hspace{2pt}%
  \begin{subfigure}[b]{0.119\linewidth}
    \includegraphics[width=\linewidth]{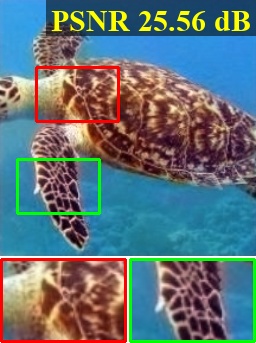}
  \end{subfigure}\hspace{2pt}%
  \begin{subfigure}[b]{0.119\linewidth}
    \includegraphics[width=\linewidth]{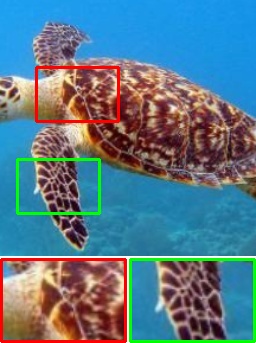}
  \end{subfigure}

  \begin{subfigure}[b]{0.119\linewidth}
    \includegraphics[width=\linewidth]{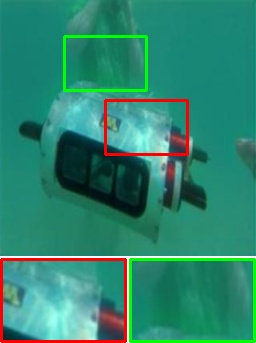}
  \end{subfigure}\hspace{2pt}%
  \begin{subfigure}[b]{0.119\linewidth}
    \includegraphics[width=\linewidth]{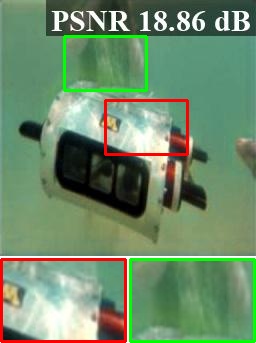}
  \end{subfigure}\hspace{2pt}%
  \begin{subfigure}[b]{0.119\linewidth}
    \includegraphics[width=\linewidth]{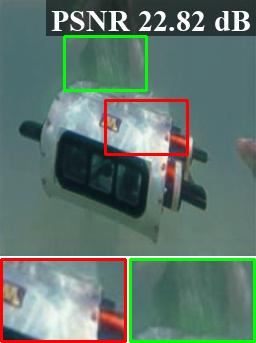}
  \end{subfigure}\hspace{2pt}%
  \begin{subfigure}[b]{0.119\linewidth}
    \includegraphics[width=\linewidth]{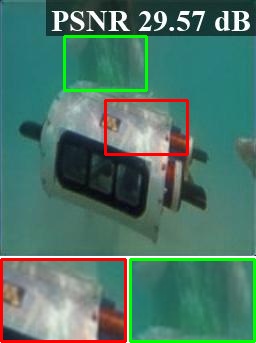}
  \end{subfigure}\hspace{2pt}%
  \begin{subfigure}[b]{0.119\linewidth}
    \includegraphics[width=\linewidth]{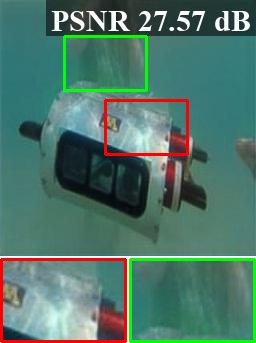}
  \end{subfigure}\hspace{2pt}%
  \begin{subfigure}[b]{0.119\linewidth}
    \includegraphics[width=\linewidth]{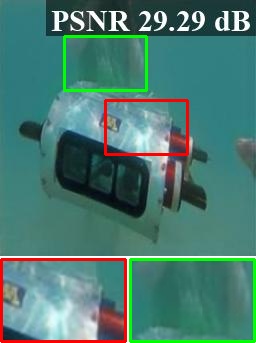}
  \end{subfigure}\hspace{2pt}%
  \begin{subfigure}[b]{0.119\linewidth}
    \includegraphics[width=\linewidth]{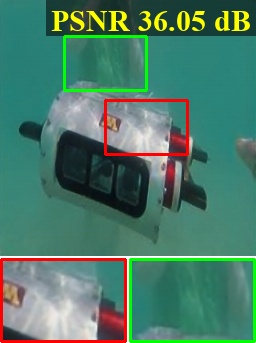}
  \end{subfigure}\hspace{2pt}%
  \begin{subfigure}[b]{0.119\linewidth}
    \includegraphics[width=\linewidth]{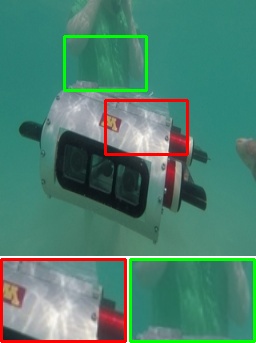}
  \end{subfigure}

  \begin{subfigure}[b]{0.119\linewidth}
    \includegraphics[width=\linewidth]{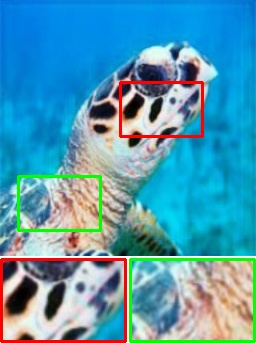}
  \end{subfigure}\hspace{2pt}%
  \begin{subfigure}[b]{0.119\linewidth}
    \includegraphics[width=\linewidth]{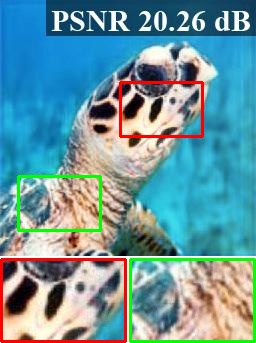}
  \end{subfigure}\hspace{2pt}%
  \begin{subfigure}[b]{0.119\linewidth}
    \includegraphics[width=\linewidth]{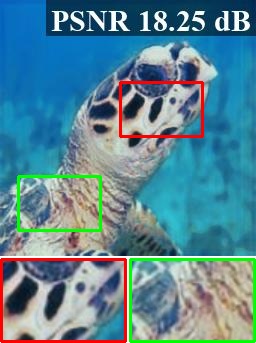}
  \end{subfigure}\hspace{2pt}%
  \begin{subfigure}[b]{0.119\linewidth}
    \includegraphics[width=\linewidth]{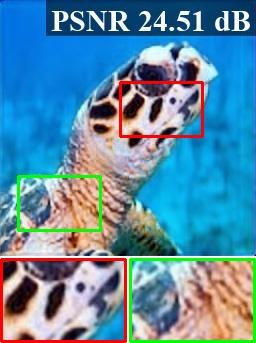}
  \end{subfigure}\hspace{2pt}%
  \begin{subfigure}[b]{0.119\linewidth}
    \includegraphics[width=\linewidth]{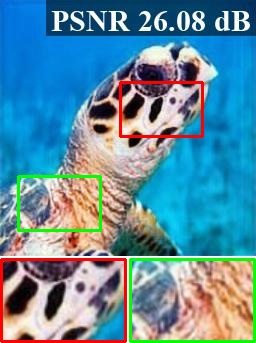}
  \end{subfigure}\hspace{2pt}%
  \begin{subfigure}[b]{0.119\linewidth}
    \includegraphics[width=\linewidth]{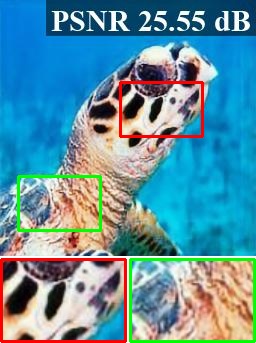}
  \end{subfigure}\hspace{2pt}%
  \begin{subfigure}[b]{0.119\linewidth}
    \includegraphics[width=\linewidth]{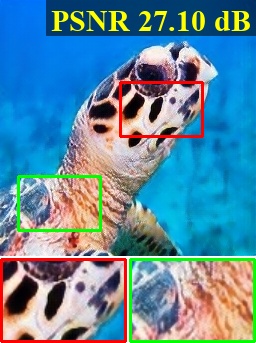}
  \end{subfigure}\hspace{2pt}%
  \begin{subfigure}[b]{0.119\linewidth}
    \includegraphics[width=\linewidth]{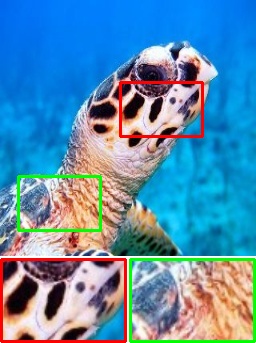}
  \end{subfigure}

  \begin{subfigure}[b]{0.119\linewidth}
    \includegraphics[width=\linewidth]{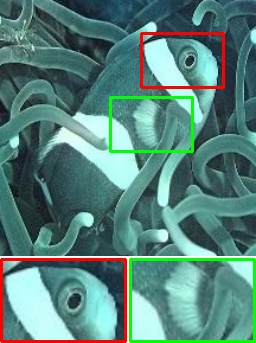}
    \caption*{\scriptsize (a) Input}
  \end{subfigure}\hspace{2pt}%
  \begin{subfigure}[b]{0.119\linewidth}
    \includegraphics[width=\linewidth]{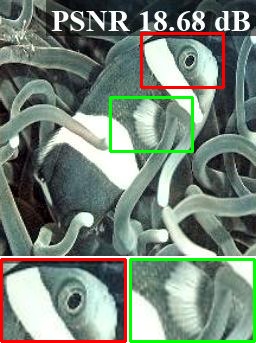}
    \caption*{\scriptsize (b) WD}
  \end{subfigure}\hspace{2pt}%
  \begin{subfigure}[b]{0.119\linewidth}
    \includegraphics[width=\linewidth]{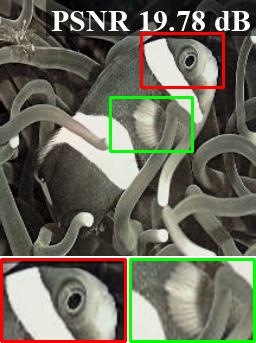}
    \caption*{\scriptsize (c) USUIR}
  \end{subfigure}\hspace{2pt}%
  \begin{subfigure}[b]{0.119\linewidth}
    \includegraphics[width=\linewidth]{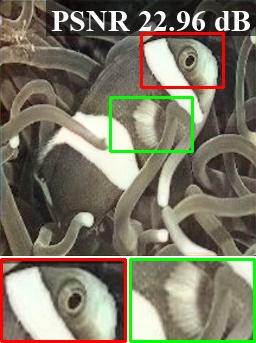}
    \caption*{\scriptsize (d) UTrans}
  \end{subfigure}\hspace{2pt}%
  \begin{subfigure}[b]{0.119\linewidth}
    \includegraphics[width=\linewidth]{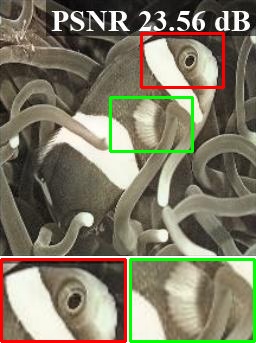}
    \caption*{\scriptsize (e) URSCT}
  \end{subfigure}\hspace{2pt}%
  \begin{subfigure}[b]{0.119\linewidth}
    \includegraphics[width=\linewidth]{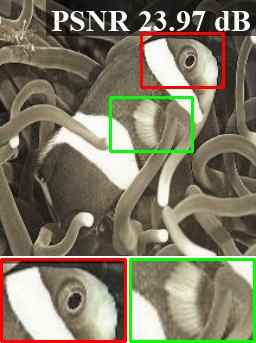}
    \caption*{\scriptsize (f) SGNet}
  \end{subfigure}\hspace{2pt}%
  \begin{subfigure}[b]{0.119\linewidth}
    \includegraphics[width=\linewidth]{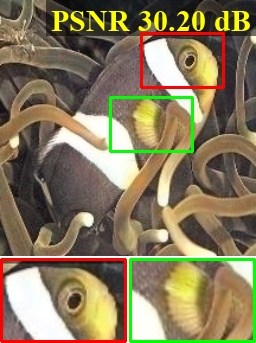}
    \caption*{\scriptsize (g) Ours}
  \end{subfigure}\hspace{2pt}%
  \begin{subfigure}[b]{0.119\linewidth}
    \includegraphics[width=\linewidth]{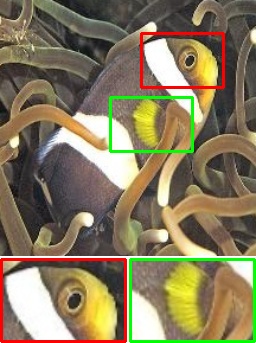}
    \caption*{\scriptsize (h) GT}
  \end{subfigure}
  \caption{Visual comparison of enhancement results sampled from the EUVP dataset. 
  }
  \label{fig:vr}
\end{figure}

We also present qualitative visual comparisons illustrated in \Cref{fig:vr}, which show that our method yields perceptually greater enhancements. Unlike other approaches, which often produce color distortions, blurry textures, or overly smoothed results, our images are more naturally colored, have enhanced sharpness, improved contrast, and increased vibrancy. The highlighted red and green regions in the images clearly illustrate our method's capability to restore intricate textures and subtle color nuances effectively. These findings emphasize our method's robustness across diverse underwater conditions, including both brightly illuminated and shadowed areas.

Collectively, both quantitative and qualitative results validate the superiority and effectiveness of our proposed method in underwater image enhancement tasks.

\begin{table}[ht]
    \centering
    \caption{Quantitative results for ablation studies in EUVP dataset. The best is marked in bold.}
    \label{tab:ablation}
    \begin{tabular}{lcccc|ccccc}
        \toprule
        Settings & BL & ViT & FAT & FAST & PSNR\textuparrow & SSIM\textuparrow & LPIPS\textdownarrow & UCIQE\textuparrow & $\Delta$ E\textdownarrow \\
        \midrule
        (a) & \ding{51} & \ding{55} & \ding{55} & \ding{55} & 21.145 & 0.911 & 0.308 & 0.807 & 13.603 \\
        (b) & \ding{51} & \ding{51} & \ding{55} & \ding{55} & 22.759 & 0.914 & 0.264 & 0.820 & 8.425 \\
        (c) & \ding{51} & \ding{55} & \ding{51} & \ding{55} & 24.543 & 0.924 & 0.220 & 0.827 & 7.151 \\
        (d) & \ding{51} & \ding{55} & \ding{55} & \ding{51} & 24.852 & 0.926 & 0.215 & 0.830 & 6.800\\
        (e) & \ding{51} & \ding{55} & \ding{51} & \ding{51} & \textbf{27.451} & \textbf{0.936} & \textbf{0.163} & \textbf{0.863} & \textbf{4.952} \\
        \bottomrule
    \end{tabular}
\end{table}

\begin{figure}[ht]
  \centering
  \begin{subfigure}[b]{0.13\linewidth}
    \includegraphics[width=\linewidth]{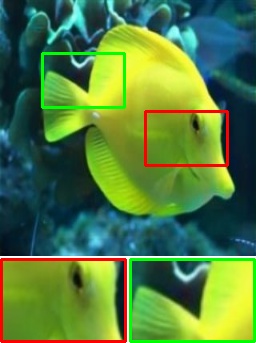}
  \end{subfigure}\hspace{2pt}%
  \begin{subfigure}[b]{0.13\linewidth}
    \includegraphics[width=\linewidth]{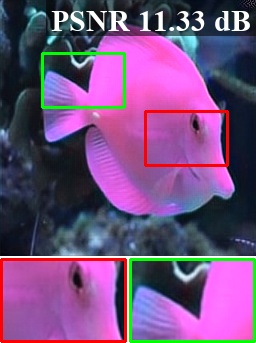}
  \end{subfigure}\hspace{2pt}%
  \begin{subfigure}[b]{0.13\linewidth}
    \includegraphics[width=\linewidth]{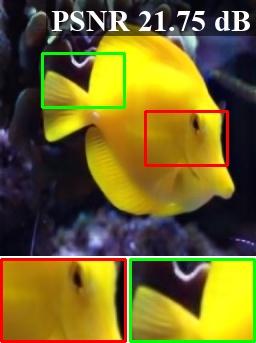}
  \end{subfigure}\hspace{2pt}%
  \begin{subfigure}[b]{0.13\linewidth}
    \includegraphics[width=\linewidth]{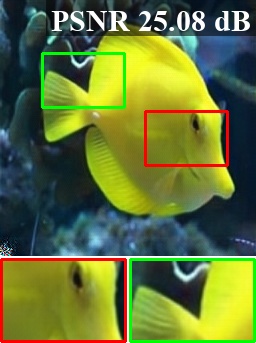}
  \end{subfigure}\hspace{2pt}%
  \begin{subfigure}[b]{0.13\linewidth}
    \includegraphics[width=\linewidth]{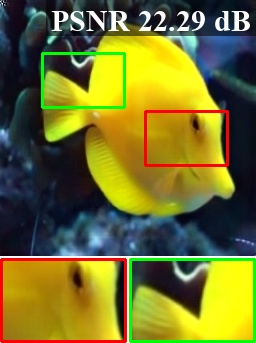}
  \end{subfigure}\hspace{2pt}%
  \begin{subfigure}[b]{0.13\linewidth}
    \includegraphics[width=\linewidth]{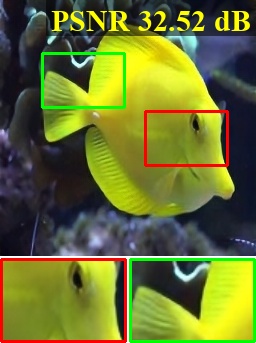}
  \end{subfigure}\hspace{2pt}%
  \begin{subfigure}[b]{0.13\linewidth}
    \includegraphics[width=\linewidth]{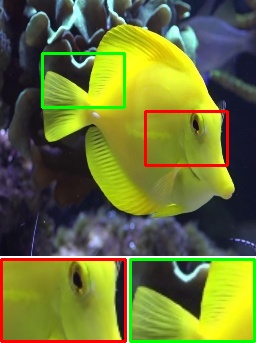}
  \end{subfigure}\hspace{2pt}%

  \begin{subfigure}[b]{0.13\linewidth}
    \includegraphics[width=\linewidth]{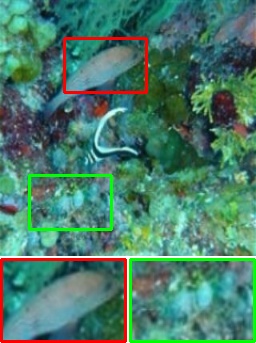}
    \caption*{\scriptsize Input}
  \end{subfigure}\hspace{2pt}%
  \begin{subfigure}[b]{0.13\linewidth}
    \includegraphics[width=\linewidth]{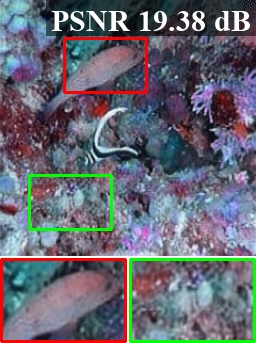}
    \caption*{\scriptsize(a) B}
  \end{subfigure}\hspace{2pt}%
  \begin{subfigure}[b]{0.13\linewidth}
    \includegraphics[width=\linewidth]{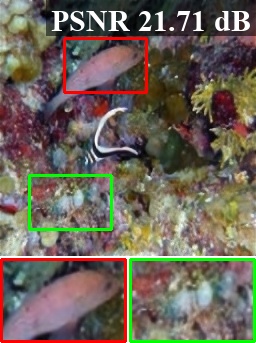}
    \caption*{\scriptsize(b) B+ViT}
  \end{subfigure}\hspace{2pt}%
  \begin{subfigure}[b]{0.13\linewidth}
    \includegraphics[width=\linewidth]{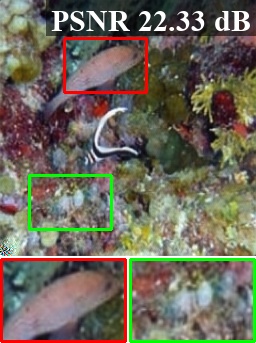}
    \caption*{\scriptsize(c) B+FAT}
  \end{subfigure}\hspace{2pt}%
  \begin{subfigure}[b]{0.13\linewidth}
    \includegraphics[width=\linewidth]{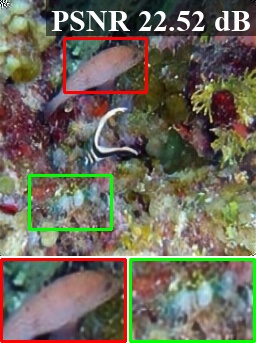}
    \caption*{\scriptsize(d) B+FAST}
  \end{subfigure}\hspace{2pt}%
  \begin{subfigure}[b]{0.13\linewidth}
    \includegraphics[width=\linewidth]{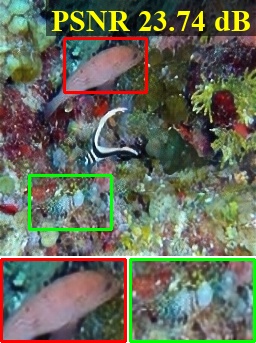}
    \caption*{\scriptsize(e) Full}
  \end{subfigure}\hspace{2pt}%
  \begin{subfigure}[b]{0.13\linewidth}
    \includegraphics[width=\linewidth]{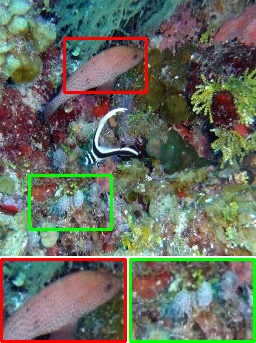}
    \caption*{\scriptsize GT}
  \end{subfigure}

  \caption{The ablation comparisons sampled from EUVP dataset. 
  }
  \label{fig:abla}
\end{figure}

\subsection{Ablation Study}

An ablation study, summarized in \cref{tab:ablation}, systematically examines the contribution of each module within our architecture. We evaluate configurations including a UNet baseline, the addition of FAT and FAST modules. We also add a simple Transformer-based bottleneck without AFFN and cross-attention.

Individual module contributions result in significant improvements in PSNR, demonstrating their effectiveness in enhancing model performance. The complete architecture, incorporating all proposed modules, yields the best overall performance, highlighting the complementary effects of each component.

Representative visual comparisons in \cref{fig:abla} further demonstrate the necessity and efficacy of the full architecture. Partial configurations reveal issues such as blur artifacts and color distortions, whereas the fully integrated model consistently provides optimal visual quality in terms of contrast, detail preservation, and color accuracy.

These findings underscore that each component significantly contributes to the model's superior performance, reinforcing the robustness and precision of our methodological framework.

\section{Conclusion}
We introduced \textbf{SFormer}, an SNR-guided framework for underwater image enhancement focusing on frequency rather than conventional spatial processing. Our FAST module operates in the Fourier domain, coupling amplitude-phase interactions with SNR priors to enhance spectral components and suppress noise. It restores texture and corrects color in diverse water conditions. The FAT bottleneck incorporates hierarchical attention to merge the low-frequency global context with the high-frequency details, ensuring balanced enhancement. Experiments show that SFormer outperforms existing CNN and Transformer methods in metrics and visual quality. Future work will explore domain-adaptive variants for turbid and deep-sea environments.

%
%
%

\begin{credits}
\subsubsection{\ackname} This work was supported in part by Shenzhen Polytechnic University Research Fund (Grant No. 6025310023K) and the National Natural Science Foundation of China (Grant No. 62272313), in part by the Science and Technology Development Fund, Macau SAR, under  Grant 0141/2023/RIA2 and 0193/2023/RIA3, and the University of Macau under Grant MYRG-GRG2024-00065-FST-UMDF.

\end{credits}
\bibliographystyle{splncs04}
\bibliography{ref}

\begin{thebibliography}{10}
\providecommand{\url}[1]{\texttt{#1}}
\providecommand{\urlprefix}{URL }
\providecommand{\doi}[1]{https://doi.org/#1}

\bibitem{akkaynak2018revised}
Akkaynak, D., Treibitz, T.: A revised underwater image formation model. In:
  CVPR. pp. 6723--6732 (2018)

\bibitem{akkaynak2019sea}
Akkaynak, D., Treibitz, T.: Sea-thru: A method for removing water from
  underwater images. In: CVPR. pp. 1682--1691 (2019)

\bibitem{ancuti2017color}
Ancuti, C.O., Ancuti, C., De~Vleeschouwer, C., Bekaert, P.: Color balance and
  fusion for underwater image enhancement. TIP  \textbf{27}(1),  379--393
  (2017)

\bibitem{ancuti2012enhancing}
Ancuti, C., Ancuti, C.O., Haber, T., Bekaert, P.: Enhancing underwater images
  and videos by fusion. In: CVPR. pp. 81--88 (2012)

\bibitem{chen2024uwformer}
Chen, W., Lei, Y., Luo, S., Zhou, Z., Li, M., Pun, C.M.: Uwformer: Underwater
  image enhancement via a semi-supervised multi-scale transformer. In: IJCNN.
  pp.~1--8 (2024)

\bibitem{cong2023pugan}
Cong, R., Yang, W., Zhang, W., Li, C., Guo, C.L., Huang, Q., Kwong, S.: Pugan:
  Physical model-guided underwater image enhancement using gan with
  dual-discriminators. TIP  \textbf{32},  4472--4485 (2023)

\bibitem{drews2013transmission}
Drews, P., Nascimento, E., Moraes, F., Botelho, S., Campos, M.: Transmission
  estimation in underwater single images. In: ICCV Workshops. pp. 825--830
  (2013)

\bibitem{fabbri2018enhancing}
Fabbri, C., Islam, M.J., Sattar, J.: Enhancing underwater imagery using
  generative adversarial networks. In: ICRA. pp. 7159--7165 (2018)

\bibitem{fu2022unsupervised}
Fu, Z., Lin, H., Yang, Y., Chai, S., Sun, L., Huang, Y., Ding, X.: Unsupervised
  underwater image restoration: From a homology perspective. In: AAAI. vol.~36,
  pp. 643--651 (2022)

\bibitem{ghani2017automatic}
Ghani, A.S.A., Isa, N.A.M.: Automatic system for improving underwater image
  contrast and color through recursive adaptive histogram modification.
  Computers and Electronics in Agriculture  \textbf{141},  181--195 (2017)

\bibitem{guo2025underwater}
Guo, X., Chen, X., Wang, S., Pun, C.M.: Underwater image restoration through a
  prior guided hybrid sense approach and extensive benchmark analysis. TCSVT
  (2025)

\bibitem{he2010single}
He, K., Sun, J., Tang, X.: Single image haze removal using dark channel prior.
  TPAMI  \textbf{33}(12),  2341--2353 (2010)

\bibitem{huang2023contrastive}
Huang, S., Wang, K., Liu, H., Chen, J., Li, Y.: Contrastive semi-supervised
  learning for underwater image restoration via reliable bank. In: CVPR. pp.
  18145--18155 (2023)

\bibitem{islam2020fast}
Islam, M.J., Xia, Y., Sattar, J.: Fast underwater image enhancement for
  improved visual perception. RA-L  \textbf{5}(2),  3227--3234 (2020)

\bibitem{jiang2020geometry}
Jiang, H., Li, S., Liu, W., Zheng, H., Liu, J., Zhang, Y.: Geometry-aware cell
  detection with deep learning. Msystems  \textbf{5}(1),  10--1128 (2020)

\bibitem{jiang2022underwater}
Jiang, Q., Gu, Y., Li, C., Cong, R., Shao, F.: Underwater image enhancement
  quality evaluation: Benchmark dataset and objective metric. TCSVT
  \textbf{32}(9),  5959--5974 (2022)

\bibitem{le2024medical}
Le, Z., Li, Q., Chen, H., Cai, S., Xiong, X., Huang, L.: Medical image
  encryption system based on a simultaneous permutation and diffusion framework
  utilizing a new chaotic map. Physica Scripta  \textbf{99}(5),  055249 (2024)

\bibitem{lei2025cmamrnet}
Lei, Y., Yi, F., Dong, Y., Liu, W., Zhang, X., Li, Z., Pun, C.M., Chen, X.:
  Cmamrnet: A contextual mask-aware network enhancing mural restoration through
  comprehensive mask guidance. In: BMVC (2025)

\bibitem{lei2024uie}
Lei, Y., Yu, J., Dong, Y., Gong, C., Zhou, Z., Pun, C.M.: Uie-unfold: Deep
  unfolding network with color priors and vision transformer for underwater
  image enhancement. In: DSAA. pp. 1--10 (2024)

\bibitem{li2021underwater}
Li, C., Anwar, S., Hou, J., Cong, R., Guo, C., Ren, W.: Underwater image
  enhancement via medium transmission-guided multi-color space embedding. TIP
  \textbf{30},  4985--5000 (2021)

\bibitem{li2020underwater}
Li, C., Anwar, S., Porikli, F.: Underwater scene prior inspired deep underwater
  image and video enhancement. Pattern Recognition  \textbf{98},  107038 (2020)

\bibitem{li2019underwater}
Li, C., Guo, C., Ren, W., Cong, R., Hou, J., Kwong, S., Tao, D.: An underwater
  image enhancement benchmark dataset and beyond. TIP  \textbf{29},  4376--4389
  (2019)

\bibitem{li2022few}
Li, H., Ge, S., Gao, C., Gao, H.: Few-shot object detection via high-and-low
  resolution representation. Computers and Electrical Engineering
  \textbf{104},  108438 (2022)

\bibitem{li2022monocular}
Li, H., Pun, C.M.: Monocular robust 3d human localization by global and
  body-parts depth awareness. TCSVT  \textbf{32}(11),  7692--7705 (2022)

\bibitem{li2023cee}
Li, H., Pun, C.M.: Cee-net: complementary end-to-end network for 3d human pose
  generation and estimation. In: AAAI. vol.~37, pp. 1305--1313 (2023)

\bibitem{li2021hybrid}
Li, H., Pun, C.M., Xu, F., Pan, L., Zong, R., Gao, H., Lu, H.: A hybrid feature
  selection algorithm based on a discrete artificial bee colony for parkinson's
  diagnosis. ACM Transactions on Internet Technology  \textbf{21}(3),  1--22
  (2021)

\bibitem{li2025adaptive}
Li, H., Zheng, F., Liu, Y., Xiong, J., Zhang, W., Hu, H., Gao, H.: Adaptive
  skeleton prompt tuning for cross-dataset 3d human pose estimation. In:
  ICASSP. pp.~1--5 (2025)

\bibitem{li2022beyond}
Li, K., Wu, L., Qi, Q., Liu, W., Gao, X., Zhou, L., Song, D.: Beyond single
  reference for training: Underwater image enhancement via comparative
  learning. TCSVT  \textbf{33}(6),  2561--2576 (2022)

\bibitem{li2025dppad}
Li, Q., Li, Q., Ling, B.W.K., Pun, C.M., Huang, G., Yuan, X., Zhong, G.,
  Ayouni, S., Chen, J.: Dppad-ie: Dynamic polyhedra permutating and arnold
  diffusing medical image encryption using 2d cross gaussian hyperchaotic map.
  IEEE Transactions on Consumer Electronics  (2025)

\bibitem{li2025gre}
Li, Q., Li, W., Zheng, X., Zhou, J., Zhong, W., Chen, X., Long, C.: Gre 2-mdcl:
  Graph representation embedding enhanced via multidimensional contrastive
  learning. IEEE Access  (2025)

\bibitem{li2024cross}
Li, X., Huang, G., Cheng, L., Zhong, G., Liu, W., Chen, X., Cai, M.:
  Cross-domain visual prompting with spatial proximity knowledge distillation
  for histological image classification. Journal of Biomedical Informatics
  \textbf{158},  104728 (2024)

\bibitem{liu2022twin}
Liu, R., Jiang, Z., Yang, S., Fan, X.: Twin adversarial contrastive learning
  for underwater image enhancement and beyond. TIP  \textbf{31},  4922--4936
  (2022)

\bibitem{liu2024dh}
Liu, W., Cun, X., Pun, C.M.: Dh-gan: Image manipulation localization via a dual
  homology-aware generative adversarial network. PR p. 110658 (2024)

\bibitem{liu2023coordfill}
Liu, W., Cun, X., Pun, C.M., Xia, M., Zhang, Y., Wang, J.: Coordfill: Efficient
  high-resolution image inpainting via parameterized coordinate querying. In:
  AAAI. vol.~37, pp. 1746--1754 (2023)

\bibitem{liu2024depth}
Liu, W., Shen, X., Li, H., Bi, X., Liu, B., Pun, C.M., Cun, X.: Depth-aware
  test-time training for zero-shot video object segmentation. In: CVPR. pp.
  19218--19227 (2024)

\bibitem{liu2023explicit}
Liu, W., Shen, X., Pun, C.M., Cun, X.: Explicit visual prompting for low-level
  structure segmentations. In: CVPR. pp. 19434--19445 (2023)

\bibitem{liu2024forgeryttt}
Liu, W., Shen, X., Pun, C.M., Cun, X.: Forgeryttt: Zero-shot image manipulation
  localization with test-time training. arXiv  (2024)

\bibitem{ma2022wavelet}
Ma, Z., Oh, C.: A wavelet-based dual-stream network for underwater image
  enhancement. In: ICASSP. pp. 2769--2773 (2022)

\bibitem{miao2021underwater}
Miao, Y., Zakharov, Y.V., Sun, H., Li, J., Wang, J.: Underwater acoustic signal
  classification based on sparse time--frequency representation and deep
  learning. IEEE Journal of Oceanic Engineering  \textbf{46}(3),  952--962
  (2021)

\bibitem{mu2023generalized}
Mu, P., Xu, H., Liu, Z., Wang, Z., Chan, S., Bai, C.: A generalized
  physical-knowledge-guided dynamic model for underwater image enhancement. In:
  ACM MM. pp. 7111--7120 (2023)

\bibitem{panetta2015human}
Panetta, K., Gao, C., Agaian, S.: Human-visual-system-inspired underwater image
  quality measures. IEEE Journal of Oceanic Engineering  \textbf{41}(3),
  541--551 (2015)

\bibitem{peng2023u}
Peng, L., Zhu, C., Bian, L.: U-shape transformer for underwater image
  enhancement. TIP  \textbf{32},  3066--3079 (2023)

\bibitem{peng2017underwater}
Peng, Y.T., Cosman, P.C.: Underwater image restoration based on image
  blurriness and light absorption. TIP  \textbf{26}(4),  1579--1594 (2017)

\bibitem{ren2022reinforced}
Ren, T., Xu, H., Jiang, G., Yu, M., Zhang, X., Wang, B., Luo, T.: Reinforced
  swin-convs transformer for simultaneous underwater sensing scene image
  enhancement and super-resolution. TGRS  \textbf{60},  1--16 (2022)

\bibitem{schechner2005recovery}
Schechner, Y.Y., Karpel, N.: Recovery of underwater visibility and structure by
  polarization analysis. IEEE Journal of Oceanic Engineering  \textbf{30}(3),
  570--587 (2005)

\bibitem{wang2022semantic}
Wang, D., Ma, L., Liu, R., Fan, X.: Semantic-aware texture-structure feature
  collaboration for underwater image enhancement. In: ICRA. pp. 4592--4598
  (2022)

\bibitem{wang2024novel}
Wang, J., Deng, Z., Lin, T., Li, W., Ling, S.: A novel prompt tuning for graph
  transformers: Tailoring prompts to graph topologies. In: Proceedings of the
  30th ACM SIGKDD Conference on Knowledge Discovery and Data Mining. pp.
  3116--3127 (2024)

\bibitem{wang2024beyond}
Wang, J., Deng, Z., Lin, T., Li, W., Ling, S., Lin, J.: Beyond direct
  relationships: Exploring multi-order label pair dependencies for knowledge
  distillation. In: Proceedings of the 32nd ACM International Conference on
  Multimedia. pp. 8527--8535 (2024)

\bibitem{wang2025structure}
Wang, J., Huang, G., Yuan, X., Zhong, G., Lin, T., Pun, C.M., Xie, F.: The
  structure-sharing hypergraph reasoning attention module for cnns. Expert
  Systems with Applications  \textbf{259},  125240 (2025)

\bibitem{wang2023qgd}
Wang, J., Huang, G., Zhong, G., Yuan, X., Pun, C.M., Deng, J.: Qgd-net: a
  lightweight model utilizing pixels of affinity in feature layer for
  dermoscopic lesion segmentation. IEEE Journal of Biomedical and Health
  Informatics  \textbf{27}(12),  5982--5993 (2023)

\bibitem{wang2024novel2}
Wang, J., Huang, G., Zhong, G., Yuan, X., Pun, C.M., Wang, J., Liu, J.: A novel
  hypercomplex graph convolution refining mechanism. IEEE Transactions on
  Emerging Topics in Computational Intelligence  (2024)

\bibitem{wang2023ultra}
Wang, T., Zhang, K., Shen, T., Luo, W., Stenger, B., Lu, T.:
  Ultra-high-definition low-light image enhancement: A benchmark and
  transformer-based method. In: AAAI. vol.~37, pp. 2654--2662 (2023)

\bibitem{wang2023domain}
Wang, Z., Shen, L., Xu, M., Yu, M., Wang, K., Lin, Y.: Domain adaptation for
  underwater image enhancement. TIP  \textbf{32},  1442--1457 (2023)

\bibitem{zhang6}
Wei, J., Zhang, X.: Dopra: Decoding over-accumulation penalization and
  re-allocation in specific weighting layer. ACM MM  (2024)

\bibitem{wen2023syreanet}
Wen, J., Cui, J., Zhao, Z., Yan, R., Gao, Z., Dou, L., Chen, B.M.: Syreanet: A
  physically guided underwater image enhancement framework integrating
  synthetic and real images. In: ICRA. pp. 5177--5183 (2023)

\bibitem{wu2025generative}
Wu, W., Dai, T., Chen, Z., Huang, X., Ma, F., Xiao, J.: Generative prompt
  controlled diffusion for weakly supervised semantic segmentation.
  Neurocomputing p. 130103 (2025)

\bibitem{wu2024top}
Wu, W., Dai, T., Huang, X., Ma, F., Xiao, J.: Top-k pooling with patch
  contrastive learning for weakly-supervised semantic segmentation. In: 2024
  IEEE International Conference on Systems, Man, and Cybernetics (SMC). pp.
  5270--5275 (2024)

\bibitem{wu2025image}
Wu, W., Qiu, X., Song, S., Chen, Z., Huang, X., Ma, F., Xiao, J.: Image
  augmentation agent for weakly supervised semantic segmentation.
  Neurocomputing p. 131314 (2025)

\bibitem{wu2025prompt}
Wu, W., Qiu, X., Song, S., Chen, Z., Huang, X., Ma, F., Xiao, J.: Prompt
  categories cluster for weakly supervised semantic segmentation. In:
  Proceedings of the Computer Vision and Pattern Recognition Conference. pp.
  3198--3207 (2025)

\bibitem{wuimgfu}
Wu, W., Song, S., Qiu, X., Huang, X., Ma, F., Xiao, J.: Image fusion for
  cross-domain sequential recommendation. In: Companion Proceedings of the ACM
  Web Conference 2025 (2025)

\bibitem{xu2024plaintext}
Xu, J., Liu, K., Huang, Q., Li, Q., Huang, L.: A plaintext-related and
  ciphertext feedback mechanism for medical image encryption based on a new
  one-dimensional chaotic system. Physica Scripta  \textbf{99}(12),  125220
  (2024)

\bibitem{xu2022snr}
Xu, X., Wang, R., Fu, C.W., Jia, J.: Snr-aware low-light image enhancement. In:
  CVPR. pp. 17714--17724 (2022)

\bibitem{zhang3}
Xu, Z., Zhang, X., Chen, W., Liu, J., Xu, T., Wang, Z.: Muraldiff: Diffusion
  for ancient murals restoration on large-scale pre-training. TETCI  (2024)

\bibitem{yang2015underwater}
Yang, M., Sowmya, A.: An underwater color image quality evaluation metric. TIP
  \textbf{24}(12),  6062--6071 (2015)

\bibitem{zhang2022correction}
Zhang, C., Jiang, H., Liu, W., Li, J., Tang, S., Juhas, M., Zhang, Y.:
  Correction of out-of-focus microscopic images by deep learning. Computational
  and Structural Biotechnology Journal  \textbf{20},  1957--1966 (2022)

\bibitem{zhang12}
Zhang, K., Chen, X., Zhang, X.: Adatoken-3d: Dynamic spatial gating for
  efficient 3d large multimodal-models reasoning. IROS  (2025)

\bibitem{zhang2018unreasonable}
Zhang, R., Isola, P., Efros, A.A., Shechtman, E., Wang, O.: The unreasonable
  effectiveness of deep features as a perceptual metric. In: CVPR. pp. 586--595
  (2018)

\bibitem{zhang1}
Zhang, X., Chen, F., Wang, C., Tao, M., Jiang, G.P.: Sienet: Siamese expansion
  network for image extrapolation. IEEE Signal Processing Letters  \textbf{27},
   1590--1594 (2020)

\bibitem{zhang7}
Zhang, X., Quan, Y., Gu, C., Shen, C., Yuan, X., Yan, S., Cheng, H., Wu, K.,
  Ye, J.: Seeing clearly by layer two: Enhancing attention heads to alleviate
  hallucination in lvlms. EMNLP  (2025)

\bibitem{zhang5}
Zhang, X., Shen, C., Yuan, X., Yan, S., Xie, L., Wang, W., Gu, C., Tang, H.,
  Ye, J.: From redundancy to relevance: Enhancing explainability in multimodal
  large language models. NAACL  (2024)

\bibitem{zhang10}
Zhang, X., Xu, Z., Tang, H., Gu, C., Chen, W., El~Saddik, A.: Wakeup-darkness:
  When multimodal meets unsupervised low-light image enhancement. TOMM  (2025)

\bibitem{zhang4}
Zhang, X., Xu, Z., Tang, H., Gu, C., Zhu, S., Guan, X.: Shadclips: When
  parameter-efficient fine-tuning with multimodal meets shadow removal  (2024)

\bibitem{zhang8}
Zhang, X., Zeng, F., Gu, C.: Simignore: Exploring and enhancing multimodal
  large model complex reasoning via similarity computation. Neural Networks p.
  107059 (2024)

\bibitem{zhang9}
Zhang, X., Zeng, F., Quan, Y., Hui, Z., Yao, J.: Enhancing multimodal large
  language models complex reason via similarity computation. AAAI  (2025)

\bibitem{zhang2}
Zhang, X., Zhao, Y., Gu, C., Lu, C., Zhu, S.: Spa-former: An effective and
  lightweight transformer for image shadow removal. In: IJCNN. pp.~1--8 (2023)

\bibitem{zhao2024toward}
Zhao, C., Cai, W., Dong, C., Zeng, Z.: Toward sufficient spatial-frequency
  interaction for gradient-aware underwater image enhancement. In: ICASSP. pp.
  3220--3224 (2024)

\bibitem{zhang11}
Zhao, Q., Zhang, X., Li, y., Xing, Y., Xiaosong, Y., Tang, F., Fan, S., Chen,
  X., Zhang, X., Wang, D.: Mca-llava: Manhattan causal attention for reducing
  hallucination in large vision-language models. ACM MM  (2025)

\bibitem{zhao2015deriving}
Zhao, X., Jin, T., Qu, S.: Deriving inherent optical properties from background
  color and underwater image enhancement. Ocean Engineering  \textbf{94},
  163--172 (2015)

\bibitem{zheng2024smaformer}
Zheng, F., Chen, X., Liu, W., Li, H., Lei, Y., He, J., Pun, C.M., Zhou, S.:
  Smaformer: Synergistic multi-attention transformer for medical image
  segmentation. In: BIBM (2024)

\bibitem{zheng2024lagrange}
Zheng, F., Li, Q., Li, W., Chen, X., Dong, Y., Huang, G., Pun, C.M., Zhou, S.:
  Lagrange duality and compound multi-attention transformer for semi-supervised
  medical image segmentation. arXiv  (2024)

\bibitem{zhong2025image}
Zhong, G., Chu, Y., Li, Q., Wang, T., Xu, S.: Image encryption based on 2d-cphm
  hyperchaotic map using cross-plane grouping permutation and cipher diffusion:
  G. zhong et al. Nonlinear Dynamics pp. 1--36 (2025)

\bibitem{zhou2023underwater}
Zhou, J., Liu, Q., Jiang, Q., Ren, W., Lam, K.M., Zhang, W.: Underwater camera:
  Improving visual perception via adaptive dark pixel prior and color
  correction. IJCV pp. 1--19 (2023)

\bibitem{zhu2017unpaired}
Zhu, J.Y., Park, T., Isola, P., Efros, A.A.: Unpaired image-to-image
  translation using cycle- consistent adversarial networks. In: ICCV. pp.
  2223--2232 (2017)

\bibitem{zhu2024test}
Zhu, L., Liu, W., Chen, X., Li, Z., Chen, X., Wang, Z., Pun, C.M.: Test-time
  intensity consistency adaptation for shadow detection. arXiv  (2024)

\end{thebibliography}
\end{document}